\documentclass{article}
\usepackage{ijcai23}
\usepackage{longtable}
\usepackage{times}
\usepackage{soul}
\usepackage{url}
\usepackage[hidelinks]{hyperref}
\usepackage[utf8]{inputenc}
\usepackage[small]{caption}
\usepackage{graphicx}
\usepackage{amsmath}
\usepackage{amsthm}
\usepackage{booktabs}
\usepackage{algorithm}
\usepackage{algorithmic}
\usepackage[switch]{lineno}

\usepackage{comment}

\usepackage{supertabular}
\usepackage{tabularx}

\graphicspath{ {Figures/} }

\title{Large Language Models on the Chessboard: A Study on ChatGPT's Formal Language Comprehension and Complex Reasoning Skills}

\author{
Mu-Tien Kuo$^{1,2*}$\and
Chih-Chung Hsueh$^{1,2*}$\and
Richard Tzong-Han Tsai$^{2,3}$
\affiliations
$^1$Chingshin Academy, Taiwan\\
$^2$Center of GIS, Academia Sinica\\
$^3$National Central University, Taiwan \\
\emails
\texttt{
\{11035018, 11035038\}@st.chjhs.tp.edu.tw\\
thtsai@g.ncu.edu.tw
}
}

\newcolumntype{Y}{>{\centering\arraybackslash}X}
\newcolumntype{Z}{>{\arraybackslash}p{2.5cm}}

\begin{document}

\maketitle

\begin{abstract}
    While large language models have made strides in natural language processing, their proficiency in complex reasoning tasks requiring formal language comprehension, such as chess, remains less investigated. This paper probes the performance of ChatGPT, a sophisticated language model by OpenAI in tackling such complex reasoning tasks, using chess as a case study. Through robust metrics examining both the legality and quality of moves, we assess ChatGPT's understanding of the chessboard, adherence to chess rules, and strategic decision-making abilities. Our evaluation identifies limitations within ChatGPT's attention mechanism that affect its formal language comprehension and uncovers the model's underdeveloped self-regulation abilities. Our study also reveals ChatGPT's propensity for a coherent strategy in its gameplay and a noticeable uptick in decision-making assertiveness when the model is presented with a greater volume of natural language or possesses a more lucid understanding of the state of the chessboard. These findings contribute to the growing exploration of language models' abilities beyond natural language processing, providing valuable information for future research towards models demonstrating human-like cognitive abilities. 
\end{abstract}

\def\thefootnote{*}\footnotetext{Equal contribution}\def\thefootnote{\arabic{footnote}}

\section{Introduction}
Large Language Models (LLMs) have demonstrated the ability to deliver state-of-the-art performance in few-shot and zero-shot scenarios, rapidly expanding their capabilities with very little data \cite{brown2020language,chowdhery2022palm,touvron2023llama}. Recently, the reasoning abilities of LLMs have gained notable traction within the NLP research community, as seen by the increased effort in crafting evaluation benchmarks \cite{cobbe2021training,patel2021are,shridhar2021alfworld,yang2018hotpotqa} and reason-inducing prompting strategies \cite{wei2023chainofthought,zhou2023leasttomost,wang2023selfconsistency,huang2022reasoning}. Given the growing importance of LLMs, evaluating their complex reasoning abilities in real-world applications is crucial, offering valuable insights into a wide spectrum of tasks that necessitate these abilities. In this research, we choose chess as a testing ground to evaluate ChatGPT's complex reasoning abilities. The main research question we aim to address is: How well can ChatGPT play chess, and what factors may affect its performance?

OpenAI recently released ChatGPT\footnote{\url{https://openai.com/blog/chatgpt}} (formally known as GPT-3.5), an instruction-tuned version of GPT-3 \cite{brown2020language} which shares a similar training process with InstructGPT \cite{ouyang2022training}. ChatGPT is designed to understand task intent via natural language instructions and engage in multi-prompt coherent conversations, a feature that distinguishes it from many other models. Its versatile applicability extends beyond the realm of linguistics, with uses in diverse fields where formal language application is requisite \cite{liu2023summary}, including high-stake endeavors like discovering unknown causal relationships based on observed data in the medical field \cite{tu2023causaldiscovery}. This broad usage invites an exploration into ChatGPT's capabilities to comprehend and infer formal language constructs accurately, with the results shedding light on its limitations in identical scenarios and identifying areas for improvement.

In this research, we evaluate ChatGPT's performance without additional fine-tuning. While there are other open-source LLMs available that support fine-tuning with specific task-related data, we chose ChatGPT for several reasons. Although fine-tuning could potentially enhance a model's performance in playing chess, our objective in this research is to assess the inherent reasoning capabilities and cognitive abilities of a widely-used, general-purpose LLM. Evaluating ChatGPT without any specific fine-tuning allows us to gauge its base abilities and potential limitations when faced with complex, formal tasks like chess. This provides valuable insights into the model's strengths and weaknesses, making it useful for the broader AI research community.

Chess is an ideal evaluation tool for AIs due to its easily controllable enclosed state and a well developed suite of comprehensive tools to evaluate player performance. With simple rules and adequately complicated boards, chess is a task that requires a high level of reasoning while being based on simple prior knowledge. It also provides a definitive environment in which the exact state of the board is interpretable through only formal language such as move notations. Chess is also found to be correlated to cognitive skills such as perception, memory, decision-making, and knowledge comprehension \cite{simon1988chess,gobet1998chess,burgoyne2016relationship,skills2017marcia}, providing insight for complicated real world applications that are of high stakes and also require sophisticated reasoning skills in conjunction with knowledge based decisions such as planning business strategies or medical consultation \cite{swami2013executive,obasola2022perceptions}.

In this research, we evaluate ChatGPT's cognitive abilities with chess through a chess game conversation, where after an initial prompt instructs the model to play, players exchange moves one message at a time. Through this process, we assess ChatGPT's capacity to comprehend complex scenarios and to retain information throughout the entire exchange. We define a baseline experiment that provides the minimal information required for a human to play chess (Section 2) and explore how incorporating different information and prompting strategies may affect ChatGPT's performance (Section 3). We also conduct further analysis on whether ChatGPT has a consistent strategy among its games (Section 4). Finally, we discuss ChatGPT's abilities on complex cognitive tasks, analyze the limitations of natural language trained LLMs' limitations on processing formal language and discuss whether the model has an "intent" when making moves (Section 5). Our research provides the following contributions:
\begin{enumerate}
\item We propose a diverse set of metrics that comprehensively evaluate the dimensions of ChatGPT's move validity and quality, allowing for a thorough analysis of its chess-playing abilities.
\item We evaluate how providing information in prompts or allowing the model to reason in natural language may impact ChatGPT's performance in handling tasks that require complex formal language comprehension.
\item We hypothesize on the limitations of natural language trained attention that leads to increased forgetfulness and inconsistencies in formal language, shedding light on potential areas for improvement in future LMs.
\end{enumerate}

\section{Baseline Experiment}
To evaluate ChatGPT's chess-playing abilities, we designed a baseline experiment in which we provided the minimum amount of information required for a human to play chess. Specifically, we instructed ChatGPT to play chess as the black player and provided white's first move. To ensure consistency in opponent difficulty, we chose to play against the state-of-the-art computer chess engine Stockfish 15.1, a widely recognized and powerful chess engine known for its high level of play. Since the Standard Algebraic Notation (SAN) is commonly used to communicate moves in chess, we utilized this notation to exchange moves with ChatGPT. We conducted all our experiments with the model \texttt{gpt-3.5-turbo-0301} and follow the parameters used in St\"{o}ckl~\shortcite{stockl-2021-watching} with a temperature of 1 and top-p of 0.9. A total of 1000 games were played against Stockfish, serving as the baseline for future experiments. 

\subsection{Baseline Procedure}
We initiated each game with a new chat instance and provided ChatGPT with the following prompt:

\begin{quote}
I want you to act as a rival chess player. I will start as white, and we will say our moves in reciprocal order. After my first message, I will just write my move. Please don't explain your decision and just reply with your move.

[White's first move].
\end{quote}

Since chess openings can lead to substantial variance in the subsequent moves, we aimed to exclude rare openings that could lead to edge-case games. To achieve this, we randomly selected one of the top four engine moves (e4, d4, Nf3, and e3) for each game, thereby ensuring an even opening distribution. Upon receiving the model's move, we checked whether the move was legal. If it wasn't, we regenerated the response until a legal move was provided or if 10 illegal moves were made consecutively, in which case the game was terminated. We recorded Stockfish's evaluation of the advantage of white's position and sampled a response from the top three moves provided by Stockfish. The game continued until it either met standard chess end criteria (e.g., checkmate or stalemate) or was terminated due to ten consecutive illegal moves.

\subsection{Evaluation Metrics}
In chess, generating high-quality moves is substantially more difficult than generating legal moves. Good quality moves require skills such as sophisticated board comprehension, memory, and future planning. In this study, we therefore evaluate LLMs' chess performance in two dimensions: legality and quality. Legality evaluates the model's ability to generate legal moves (i.e., moves that comply with chess rules), while quality evaluates how good a move is in terms of increasing the player's positional advantage.

Given a series of games $G$ where
\[G_i = (P_{i,1}, P_{i,2}, ..., P_{i,n_i})\]
\[P_{i,j} =
\begin{cases}
    1, & \text{Model made any number of illegal attempts}\\
    0, & \text{No illegal attempts were made}
\end{cases}
\]
a series $n$ where $n_i$ is the count of moves ChatGPT made in game $i$, a series $r$ where $r_i^j$ is the amount of illegal moves ChatGPT attempted before making a legal move on game $i$'s $j$th move, we define the metrics as follows:

We measure validity using two metrics, the Illegal Move Ratio (IMR) and Retries Before Legal Move (RBLM). IMR represents validity at the attempt level, calculating the ratio of illegal moves to total moves. Game $i$'s IMR at move $t$ is defined as follows (Game $i$'s IMR is defined as $IMR(i, n_i)$):
\[IMR(i, t)=\frac{\Sigma_{j=1}^{t}P_{i,j}}{t}\]
RBLM captures the average count of illegal moves ChatGPT makes before making a valid move. Game $i$'s RBLM at move $t$ is defined as follows (Game $i$'s RBLM is defined as $RBLM(i, n_i)$):
\[RBLM(i, t)=\frac{\Sigma_{j=1}^{t}r_i^j}{\Sigma_{j=1}^{t}P_{i,j}}\]

As making an illegal move is extremely uncommon among human chess players, we argue that there is a substantial threshold of incoherence required to make such a mistake. Therefore, IMR presents an isolated figure that only studies the distributions of these catastrophic attempts. RBLM scores are designed to represent how spread the model's next options are. A high RBLM indicates that the model has a wide range of moves deemed viable, which indicates uncertainty in the model, while low RBLM indicates a limited amount of considered moves and higher certainty (see Section 5.3 for more detail). Although one may argue that IMR and RBLM can be combined into a single metric (total illegal attempts over total attempts), this would fail to separate games that often make short bursts of illegal moves from games that suffer from a few long sequences of illegal moves. Our two-metric system enables further model motive interpretation, allowing for a more detailed analysis of why models may fail to comply with chess rules.

In assessing move quality, we employ Stockfish's advantage evaluation function, which quantifies white's positional advantage in centipawns (one hundredth of a pawn's value). A positive value signifies a favorable position for white while a negative value a favorable position for black. Given the progressive increase in evaluation throughout the course of a game (as Stockfish constantly outperforms ChatGPT), it is flawed to compute the mean Board Evaluation (BE) across all games, as experiments that utilize prompts that result in longer game durations will invariably yield higher average evaluations. Consequently, we restrict our analysis to the mean BE on the 20th move (since at least 10\% of games in each variation reach this checkpoint) within each variation. The average BE over all games can be found in Appendix A.

We also take the Games' Length (GL) into consideration. Although typically the length of games indicates very little information in chess games, most of the games ChatGPT played were terminated due to ten consecutive illegal moves. As noted in the following sections, ChatGPT's performance tends to decay as the length of the game increases, we therefore record GL to provide insight into the average required length of a game to have ChatGPT fail to generate legal moves.

In future sections, we use these metrics to evaluate ChatGPT's performance in playing chess and assess the impact of various prompting strategies on its ability to play the game.

\subsection{Baseline Performance}

The baseline experiments reveal an underwhelming performance by ChatGPT, as detailed in Table 1. ChatGPT failed to secure a win in any games and recorded a high IMR, generating an illegal move every four moves. The quality of moves, assessed by how often ChatGPT's moves improved black's advantage, was consistently poor with black rarely gaining an advantage over white. Upon observing the model's IMR and RBLM, we found that both consistently increased as the length of the games extended. This suggests that the length of games impacts the model's understanding of the board state and its adherence to the rules of chess. This observation resonates with findings from Bang \textit{et al.},~\shortcite{bang2023multitask}, where increased inconsistencies and forgetfulness were detected in prolonged conversations. We hypothesize that this trend may stem from two key aspects: ChatGPT's limited capability of retaining previous conversational context, and the model's difficulty in handling intricate game scenarios. We term the former as \textit{attention decay}, referring to the model's declining ability to reference and incorporate past conversational content into its responses over the course of an extended dialogue, and explore how prompting effects the impact of attention decay on the model.

\begin{table}

\begin{tabularx}{\columnwidth}{@{} *4{>{\centering\arraybackslash}X}@{}}
\hline
IMR   & RBLM  & GL    & BE         \\
0.26  & 6.78  & 18.79 & 253.1      \\
\hline
\end{tabularx}
\captionsetup{font=small}

\captionsetup{font=normal}
\caption{ Baseline Performance. Lower IMR and RBLM reflect better legality; higher GL signifies prolonged games; and higher BE indicates poorer move quality. }
\end{table}

\section{Incorporating Alternate Prompts}

Given ChatGPT's sub-optimal performance in move validity and high game-termination rates, we explore the potential of using prompts to enhance the model's ability to generate valid chess moves. Prior research recognizes prompting as a cost-effective method to improve LLMs' performance, sometimes even outperforming fine-tuned models \cite{reynolds2021prompt,webson-pavlick-2022-prompt,kojima2023large,wei2023chainofthought}. In this section, we aim to determine whether prompts that provide clearer instructions or assistive information can improve ChatGPT's ability to generate legal moves and impact its move quality.
To achieve this objective, we devise and implement variations of the baseline procedure employing different prompting strategies. We evaluate the effectiveness of these prompt variations by recording 400 games per variation, adhering to the baseline procedure for all steps unless specified otherwise.

\subsection{Investigating Initial Prompt Variations}

We explored the impact of altering the initial prompt in two ways. The first variation, labeled as \textbf{Int-Illegal}, involved appending the message "Please do not make illegal moves" to the original prompt. This variation tested whether ChatGPT had learned conceptual functions \cite{reynolds2021prompt} that would help it avoid illegal moves. The second variation, termed \textbf{Int-Rules}, involved including a concise summary of the rules of chess within the initial prompt. The objective of this variation was to test whether an in-prompt version of rules would increase the model's attention to generating legal moves.

\begin{table}
\begin{tabularx}{\columnwidth}{Z*{4}{Y}}
            & IMR  & RBLM & GL    & BE     \\
Baseline    & 0.26 & 6.78 & 18.79 & 253.1  \\
Int-Illegal & 0.27 & 6.86 & 18.07 & 278.6 \\
Int-Rules   & 0.33 & 7.52 & 13.15 & 364.53
\end{tabularx}
\caption{Initial Prompt Variations' Results}
\end{table}

\textbf{Results: } The Int-Illegal variation yielded results that were nearly identical to the baseline procedure. However, when asked about chess rules, ChatGPT demonstrated a perfect understanding by accurately reciting every rule multiple times. This indicates that relying solely on natural language hints is insufficient to improve model performance in chess. On the other hand, the Int-Rules variation resulted in a noticeable performance drop compared to the baseline, with a significant decrease in the average game length. We speculate that the inclusion of the rule tokens in the prompt diluted the attention received by the board state tokens, thereby compromising the model's ability to comprehend the board effectively. Our experiments revealed that ChatGPT failed to effectively utilize the provided chess rules, regardless of whether they were included as model memory or in the prompt. Furthermore, reinforcing the importance of rules did not lead to better model performance.

\subsection{Investigating Move Prompt Variations}

We investigated the effectiveness of adding information to move prompts to enhance ChatGPT's ability. To this end, we conducted two experimental variations of the baseline procedure that incorporate additional information in move prompts. 

The first variation, named \textbf{Move-Repeat}, involves appending every move made in the game to the end of the move prompt. This experiment aims to reduce the impact of ChatGPT's attention decay by increasing the appearances of tokens that date back further in the game. The second variation, termed \textbf{Move-IlgRem}, provided a reminder to ChatGPT whenever it made illegal moves. In this variation, we supplied ChatGPT with a list of its previous illegal attempts during that move and informs it that those are illegal, aiming to reduce game terminations by preventing ChatGPT from making the same mistakes repeatedly.

\textbf{Results:} The Move-Repeat variation yielded considerable improvements over the baseline in all metrics except IMR. We observed considerable enhancements in GL and BE, suggesting that Move-Repeat enables the model to generate a more concrete understanding of the board, mitigating the impact of attention decay and resulting in substantially longer games. Interestingly, the model attempts more illegal moves but requires fewer moves before reaching a legal solution. We speculate that this might be a form of model "intent," which we define as signs of the model reducing move candidates and showing more faith or determination towards a certain move. 

On the contrary, Move-IlgRem demonstrated extremely poor chess abilities. Although the model tended to avoid moves deemed illegal, the staggering RBLM suggests a strong sense of uncertainty. We hypothesize that this is due to the model's inability to differentiate game moves from moves in reminders, resulting in drastic drops in board comprehension, causing high RBLM and short games. Interestingly, Move-Repeat and Move-IlgRem variations only show effects if the information is included throughout the entire conversation. If the information is only appended after the latest move, both variations exhibit baseline-like results. This finding indicates that repetition itself may not be enough, and constant repetition might be required to achieve compelling improvements.

\begin{table}
\begin{tabularx}{\columnwidth}{Z*{4}{Y}}
            & IMR  & RBLM & GL    & BE     \\
Baseline    & 0.26 & 6.78 & 18.79 & 253.1  \\
Move-Repeat & 0.31 & 5.82 & 23.97 & 284.99 \\
Move-IlgRem & 0.23 & 9.33 & 12.96 & 314.38  \\
\end{tabularx}
\caption{Move Prompt Variations' Results. Lower IMR and RBLM reflect better legality; higher GL signifies prolonged games; and higher BE indicates poorer move quality.}
\end{table}

\begin{table}
\begin{tabular}{ l p{6.5cm} } 
\hline
\textbf{Baseline} & Move: \textit{[Stockfish's move]}\\
\hline
\textbf{Example} & Move: Nd7 \\
\hline\\
\hline
\textbf{Move-} & Move: \textit{[Stockfish's Move]},\\
\textbf{Repeat}& Previous Moves: \textit{[Previous Move]}\\
\hline
\textbf{Example} & Move: Nf6, \\
 & Previous Moves: 1. Nf3 d5 2. d4 e6 3. g3 Bd6 4. c4 c6 5. Bg2\\
\hline\\
\hline
\textbf{Move-} & Move: \textit{[Stockfish's move]} (moves \textit{[Illegal}\\
\textbf{IlgRem}&\textit{moves made]} are illegal).\\
\hline
\textbf{Example} & Move: Nd7 (moves b2, c5 are illegal). \\
\hline
\end{tabular}
\caption{All variations of the Move Prompts }
\end{table}

\subsection{Reasoning in Natural Language}

Recent work has shown that allowing LLMs to reason in natural language can substantially enhance model performance, both in a few-shot and a zero-shot manner \cite{wei2023chainofthought,zhou2023leasttomost,kojima2023large}. The improvements these methods bring are most often observed on a limited set of benchmarks, namely arithmetic, commonsense, and symbolic reasoning tasks \cite{wei2023chainofthought,zhou2023leasttomost,wang2023selfconsistency,kojima2023large}. As these benchmarks are relatively straightforward compared to chess, we tested the extent to which allowing models reasoning in natural language can improve ChatGPT's chess abilities.

To this end, we designed variations where models were encouraged to reason in natural language before making their move. Another \texttt{gpt-3.5-turbo-0301} instance was given the model's response and eight shots of examples to extract the final move in the format of the SAN notation. The sentence was not injected if the model had already learned to emulate this behavior. We conducted three experiments: \textbf{Rsn-Simple}, \textbf{Rsn-CoT}, and \textbf{Rsn-DropCoT}. In Rsn-Simple, the model was asked to "analyze the board and explain your move" and was offered the most recent analysis as an example. Rsn-CoT and Rsn-DropCoT followed Kojima \textit{et al.}~\shortcite{kojima2023large} in encouraging the model to reason with Chain of Thought (CoT) \cite{wei2023chainofthought} in a zero-shot manner. In the initial prompt, an instruction was included to "provide a step-by-step analysis", and an additional message \texttt{Let's think step by step.} was inserted before the model's explanation. We also investigated whether removing prior reasoning in the conversation affects model performance by conducting two versions of CoT: Rsn-CoT (which keeps up to eight instances of prior reasoning) and Rsn-DropCoT (which only keeps the most recent one).

\textbf{Results: } In all variations involving natural language reasoning (NL reasoning), we observed significant decrements in RBLM, indicating that NL reasoning also helps LLMs generate "intent." However, the presence of intent does not necessarily correlate with better game performance. Allowing NL reasoning significantly impaired the model's move quality, which we attribute to the excessive amount of erroneous information in the model's reasoning, analysis, and game state description. This, in turn, misled the model into formulating strategies based on model hallucinations. Interestingly, Kojima \textit{et al.},~\shortcite{kojima2023large}'s prompting strategies led to worse move legality and quality compared to Rsn-Simple. Upon close examination of the dialogues within Rsn-Simple and Rsn-DropCoT, we identify a major distinction in the amount of spurious information. Rsn-Simple responses tend to be short, containing only one to two sentences. In contrast, Rsn-DropCoT responses typically consists of an evaluation of the opponent's move, a list of plausible moves along with the ramifications of each, and a decision of what the models believes is the best course of action. This however, introduces an a considerably greater amount of erroneous information, exacerbating the model's susceptibility to illegitimate information and resulting in the decreased legality. 

Additionally, we observed the "one-shot contamination" effect discussed by Reynolds and McDonell~\shortcite{reynolds2021prompt}, where Rsn-DropCoT performed comparatively worse than Rsn-CoT. Rsn-CoT may have better performance due to it allowing models to see a more diverse set of explanations. Although most explanations are incorrect, the exposure to more diverse information may bring performance gains similar to how sampling multiple responses improves model performance in Wang \textit{et al.},~\shortcite{wang2023selfconsistency}. Intuitively, Rsn-DropCoT exhibited a more consistent strategy (indicated by lower RBLM) compared to Rsn-CoT, suggesting that having only a single response better retains consistency within the model's decision-making process. Future research should continue to explore how different types of NL reasoning impact model performance in complex tasks like chess.

\begin{table}
\begin{tabularx}{\columnwidth}{Z*{4}{Y}}
              & IMR  & RBLM & GL    & BE     \\
Baseline      & 0.26 & 6.78 & 18.79 & 253.1 \\
Rsn-Simple    & 0.34 & 5.84 & 18.11 & 412.26 \\
Rsn-CoT       & 0.37 & 5.82 & 18.45 & 492.4 \\
Rsn-DropCoT   & 0.4  & 5.31 & 19.56 & 525.34 \\
Dsc-Base      & 0.47 & 5.02 & 19.3  & 763.11 \\
\end{tabularx}
\caption{NL Reasoning \& Board Description Variations' Results. Lower IMR and RBLM reflect better legality; higher GL signifies prolonged games; and higher BE indicates poorer move quality.} 
\end{table}

\subsection{Describing State in Natural Language}

Given that ChatGPT is primarily trained on natural language data, a compelling research question arises regarding the extent to which substituting formal language with natural language can enhance ChatGPT's capacity for intricate reasoning. Therefore, we designed an experiment to investigate whether formal language is the main factor contributing to ChatGPT's unsatisfactory chess abilities. To this extent, we crafted a prompt that supplements a natural language board description on each move, providing information about each piece's location and relation. This message is appended after the phrase \texttt{"After my move, the board state is a follows: \textit{board state}"} in the move prompt. We began by describing white's state, including details about the quantity of each piece type and each piece's location and relation with other pieces in a prompt such as follows.
\begin{quote}
    White has \textit{[quantity]} \textit{[piece-type]} left.
    
    A \textit{[piece-type]} is on \textit{[square]}, can capture \textit{[targets]}, can be captured by \textit{[attackers]}, and is defended by \textit{[defenders]}.
    
    ...
\end{quote}
An additional message is added behind pawns that are an en passant target. We last specify whether white has kingside and queenside castling rights. The description, in the same format, is then repeated for black. After the description, we ask ChatGPT to make its next move. Due to input token limitations, we only retained the most recent description of the chess board in the conversation. Implementing the variation described above, we conduct the experiment \textbf{Dsc-Base}.

\textbf{Results: } The results of Dsc-Base was surprisingly underwhelming. As shown in table 5, this variation had the highest IMR and BE across all experiments, demonstrating a substantially worse chess performance. However, lower RBLM indicated a stronger decision making intent by the model. Upon inspecting the model's responses, we did not find a high volume of erroneous information like those in the NL reasoning variation. In fact, the responses instead were relatively plain and contained only the model's moves made in the SAN notation. We posit that the drop in IMR and move quality can be attributed to the model's failure to effectively apply its learned chess cognitive functions from SAN notation to the natural language board descriptions. As most chess games on the internet are recorded in the SAN notation, we propose that the conceptual functions that play chess \cite{reynolds2021prompt} in ChatGPT is more active when the input is in the SAN format instead of a much more general format (i.e., natural language). Further evidence supporting this notion is the model's consistent choice to make the move in the SAN notation without user specification. As for the lower RBLM, the model's trait of having a stronger intent when given natural language inputs remain unchanged, thus resulting in the lower RBLM value.



\section{Analyzing ChatGPT's Strategic Behavior}

We next analyze whether a consistent behavior can be observed in ChatGPT. Building upon the \textit{faithfulness} concept in evaluating NLP systems' explainability \cite{jacovi-goldberg-2020-towards,deyoung2020eraser}, we draw inspiration from Jacovi and Goldberg~\shortcite{jacovi-goldberg-2020-towards} and assess the consistency of ChatGPT's moves as an indicator of its strategic behavior.

\subsection{Illegal Move Diversity}
One of the most significant issues in ChatGPT's chess performance is its propensity for making illegal moves. The root causes of these illegal moves may be attributed to two opposed reasons. The first being that the model may resort to generating arbitrary moves due to a lack of clear direction. The other may be an exhibition of the model's strong "intent" to achieve an objective (for instance, mirroring the human thought process of capturing the opponent's high-value pieces) and, in doing so, it may overlook the rules of the game. To discern the tendency of the model's performance, we introduce the Move Repetition Score (MRS). This score quantifies the similarity between ChatGPT's illegal moves. The MRS for each game is calculated as follows:
\[MRS = \frac{\Sigma_{i=1}^n \Sigma_{j=1}^{c_i} (\frac{c_i^j}{a_i})^2 }{n} \]
where $n$ is the count of moves where ChatGPT attempted illegal moves, $a_i$ is the count of attempts of illegal moves on move $i$, $c_i$ is the count of unique illegal moves ChatGPT attempted on move $i$, and $c_i^j, 1 \leq j \leq c_i$ is the model's total attempts of the $j$th unique move on move $i$. We then calculate the average of all games' MRS to obtain each variation's MRS.

\textbf{Results: } The MRS displays considerable variation across different iterations, indicating that the model's thought process is influenced by prompting. In general, variations that involve more natural language (both in board description and model reasoning) elicit more consistent illegal moves, thereby suggesting that retaining any amount of natural language in the conversation history can enhance the model's capability to pursue a goal consistently. The elevated MRS observed in the Move-Repeat variation is noteworthy. As the Move-Repeat variation theoretically provides a more accurate board representation, we posit that allowing the model to perceive the board with fewer misrepresentations also enables it to make more strategically consistent moves. Variations without NL reasoning (i.e., Baseline and Int-Illegal) resulted in the model demonstrating more erratic attempts, but still perform better than variations that have distractions (i.e., Int-Rules and Move-IlgRem). The Int-Rules variation, which incorporates information that we find is distracting to the model, produced more arbitrary outcomes. The Move-IlgRem variation is an exception to the correlation between MRS and RBLM. Due to its design to deliberately avoid illegal move repetition, the significant drop in MRS indicates the model's attempts to avoid making the same illegal moves, but the extremely high RBLM demonstrates an exorbitant amount of randomness in the model's moves, demonstrating the severe impact the illegal move reminders have on ChatGPT's board comprehension.

\begin{table}
    \centering
    \renewcommand{\arraystretch}{1.5} 
    \begin{tabularx}{\columnwidth}{
        >{\centering\arraybackslash}X
        >{\centering\arraybackslash}X
        >{\centering\arraybackslash}X}
        \hline
        Baseline & Int-Illegal & Int-Rules \\
        0.51 & 0.5 & 0.44 \\
        \hline
        Move-Repeat & Move-IlgRem & Rsn-Simple \\
        0.64 & 0.06 & 0.63 \\
        \hline
        Rsn-CoT & Rsn-DropCoT & Dsc-Base \\
        0.59 & 0.63 & 0.6 \\
        \hline
    \end{tabularx}
    \caption{Average MRS per Variation}
\end{table}

\subsection{Game Level Performance Evaluation}

In this subsection, we aim to further investigate ChatGPT's behavior in chess games by conducting a manual analysis of the game conversations. Our analysis focuses on assessing the quality of ChatGPT's moves, insights, and suggestions to gain a deeper understanding of its chess-playing capabilities. We randomly selected 50 games from the Rsn-Simple variation and truncated 30-70\% of their moves (the percentage for each game was decided randomly). For each game, we prompted ChatGPT to simulate a skillful chess player and find black's best move. The model's response was then evaluated according to three criteria: Alignment, Insight, and Suggestions. Alignment measures whether one of ChatGPT's moves matches a move actually played in the original game. Insight measures the correctness of ChatGPT's analysis of the board (e.g., potential threats, strategies or possibilities for checkmate). Suggestions evaluates whether all of ChatGPT's suggested moves are among the top four moves recommended by Stockfish for that particular board position.

The results of our evaluation revealed that only 9 out of the 50 games exhibited proper alignment, 16 demonstrated accurate insight, and 39 had valid suggestions. The low alignment scores substantiate the significant role that model reasoning plays in the differences observed between the model's behavior in the baseline and Rsn-Simple experiments. This is attributed to ChatGPT's reasoning process in this experiment, which since ChatGPT typically makes a move first and then provides an explanation, closely mirrors that of the baseline experiments. Further enhancing this argument, little correlation was found between the acceptance of insight and suggestions (Pearson's $r=0.05$), indicating that the insight provided after moves does not influence the moves the model makes. The poor insight scores align with the observations in Section 3.3, where the model tends to hallucinate board information such as achievements and threats, resulting in incorrect strategic analysis. However, ChatGPT performed relatively better in terms of suggestions, with a significant proportion of games suggesting moves that Stockfish ranked among the best four. This is consistent with the acceptable performance observed in the baseline experiment's move quality. Overall, our manual analysis of ChatGPT's games supports our previous arguments and is consistent with our statistical calculations, highlighting the reasoning process's impact on model decisions and the model's challenges in generating accurate insights.

\section{Discussion}

\subsection{ChatGPT's Performance Overview}

Despite ChatGPT's demonstrated proficiency in natural language processing, it displays substantial limitations when it comes to playing chess. In the 3200 games played during our experiment, ChatGPT failed to secure any victories. Furthermore, only 1.59\% of games concluded naturally in accordance with the standard rules of chess. Early terminations frequently occurred at ChatGPT's second or third move, and games' IMR and BE continuously increased throughout games, illustrating its struggle to both adhere to the game's rules and navigate the increasing strategic complexity as games progressed. The average game length across all experimental variations was significantly lower than the human average of 74.28 moves per game, as documented by Deleo and Guven~\shortcite{deleo2022chesslm}. This disparity highlights the considerable gap between ChatGPT's performance and human expertise in chess.

Although GPT models like ChatGPT are trained to memorize domain-specific information, such as chess rules, our experiments reveal a clear challenge for ChatGPT in applying these rules effectively. In the context of chess, every game introduces unique strategic and positional situations, requiring a dynamic application of chess rules. The high IMR and RBLM across all variations underscore ChatGPT's difficulty in dynamically applying these memorized rules to novel, complex scenarios. This observation persisted even when clearer board representations were provided, suggesting that the high IMR may stem more from issues with rule adherence than from a lack of board state comprehension.

These findings raise critical concerns about deploying ChatGPT in high-stakes contexts that demand the accurate application of a comprehensive rule set, such as providing medical diagnostics or legal interpretations. The model's observed inability to effectively self-regulate, despite possessing an understanding of the rules, questions its reliability in such scenarios. Our study, although rooted in the context of chess, sheds light on potential limitations of ChatGPT and similar models when tasked with complex situations that necessitate formal logic and strategic planning, enabling further research to better understand and address these limitations.

\subsection{Limitations of ChatGPT's Self-Attention Mechanism in Chess Gameplay}

ChatGPT's self-attention mechanism plays a crucial role in its performance, especially in chess gameplay. Our experiments reveal two critical limitations of transformer-based LMs like ChatGPT when trained on natural language.

The first limitation is related to the increase in IMR and RBLM over the course of a game. As highlighted by \cite{stockl-2021-watching}, GPT-2 models are found to devote less attention to SAN notation tokens that are farther away from the latest input. Since a complete game memory is paramount for models to accurately track the board state \cite{toshniwal_chess_2022}, we postulate that ChatGPT's disproportionate attention allocation might be the cause of a significant portion of its mistakes. This limitation is evident across all variations that depend on formal language but don't actively reinforce the game state, which presents a challenge to the effectiveness of LLMs in tasks requiring extended conversation memory.

The second limitation pertains to the tendency of natural language trained LLMs to neglect formal language where tokens are used in an unconventional manner. Maynez \textit{et al.}~\shortcite{maynez-etal-2020-faithfulness} noted that LMs typically remain indifferent to noises or artifacts in training data, which we argue may also apply to formal languages like chess notations. This issue is particularly evident in the Int-Rules variation, where despite the introduction of helpful data, ChatGPT's performance dropped substantially. We hypothesize that this may be due to the model shifting its focus towards the rules, thereby reducing the attention allocated to the game board.

These identified limitations, while challenging, also provide valuable insights for future research. For instance, addressing the second limitation might involve frequent repetition of formal language sequences, potentially leading to more substantial improvements in game performance. Our findings is a first step towards investigating techniques such as token repetition's impact on model performance, laying the ground work for future work to explore how we can mitigate the impact of disproportionate attention allocation.

\subsection{Intent Behind LLMs' Decisions}

Do LLMs actually exhibit strategies or "intent" in their gameplay, or are they simply attempting to randomly predict legal moves? Our investigation into this issue involves the model's RBLM and MRS, which we find a striking correlation between evidenced by a Pearson correlation coefficient of $r = -0.86$. Our findings corroborate that a decreased RBLM is indicative of the model contemplating fewer moves. This, in turn, signifies a heightened degree of confidence in the selection of moves at the level of output token probability distribution. Therefore, when we observe low RBLM and high MRS, we can confidently infer that both increased natural language in the conversation and providing better board representation enhance the model's "intent." The effect of board representation is especially noteworthy, as no natural language clues were provided in these cases, making it impossible for the model to exclude moves for the purpose of maintaining a consistent narrative. However, it is important to bear in mind that "intent," as we define it here, doesn't necessarily equate to better move quality—it simply means that the model is making decisions in a non-random manner. We encourage future work to conduct detailed examination of the model's decisions across multiple moves to evaluate the presence of a consistent, long-term strategy.

\section {Conclusion}

In summary, our investigation reveals that despite its exceptional capabilities in natural language processing, ChatGPT faces considerable challenges with complex reasoning tasks involving formal language, as evidenced by its chess gameplay performance. The model's attention mechanism exhibits limitations in adequately recognizing tokens used in formal language, resulting in a suboptimal understanding of the game board. Interestingly, our findings indicate that consistent repetition of relevant information throughout a conversation can partially alleviate this limitation. Yet, despite ChatGPT's capacity to learn and internalize rules, the model struggles with self-regulation, which neither in-prompt instructions nor improved board comprehension appear to enhance. Additionally, we find that the model's decision-making focus, or "intent," can be strengthened by allowing NL reasoning, providing NL chessboard descriptions or enabling a clearer representation of the game board. Future research could examine how this disproportionate attention allocation impacts other tasks that involve formal language and necessitate complex cognitive processing. In conclusion, while ChatGPT stands as a remarkable advancement in artificial intelligence, it continues to face significant limitations, especially in non-linguistic contexts. These findings highlight the necessity for further refinement before ChatGPT, and models of its kind, can be considered reliable tools for practical applications requiring complex cognition akin to human abilities.

\section*{Acknowledgments}
We extend our acknowledgement to Mr. Cheng-Chi Lu for his mathematical consultations and astute insights, which have greatly enhanced the clarity and precision of this work. His expertise has been a valuable asset in the crafting of this paper. We would also like to express our profound gratitude to Ms. Yi-Pin Lin, whose guidance and unwavering support have been instrumental in this research. 

\bibliographystyle{named}
\bibliography{ijcai23}

\appendix

\section{Full Experiment Data}

\begin{table}[H]
    \centering
    \renewcommand{\arraystretch}{1.3} 
    \begin{tabularx}{\columnwidth}{
        >{\centering\arraybackslash}X
        >{\centering\arraybackslash}X
        >{\centering\arraybackslash}X}
        \hline
        Baseline & Int-Illegal & Int-Rules \\
        88.38        & 84.38    & 90.84 \\
        \hline
        Move-Repeat & Move-IlgRem & Rsn-Simple \\
        148.24       & 71.7      & 145.64 \\
        \hline
        Rsn-CoT & Rsn-DropCoT & Dsc-Base \\
        166.79      & 194.12   & 293.35 \\
        \hline
    \end{tabularx}
    \caption{ Average BE (Full Game) per Variation }
\end{table}

\begin{figure}[H]
    \centering
    \includegraphics[width=7.2cm]{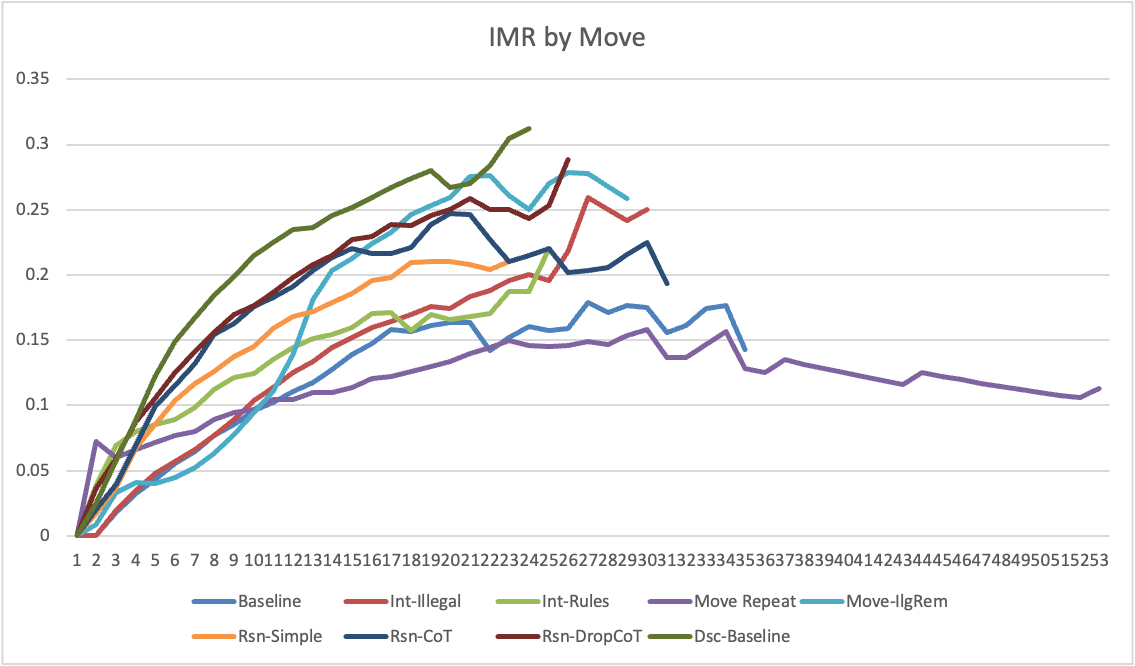}
    \caption{Average IMR by Move}
    \label{fig:apd_imr}
\end{figure}

\begin{figure}[H]
    \centering
    \includegraphics[width=7.2cm]{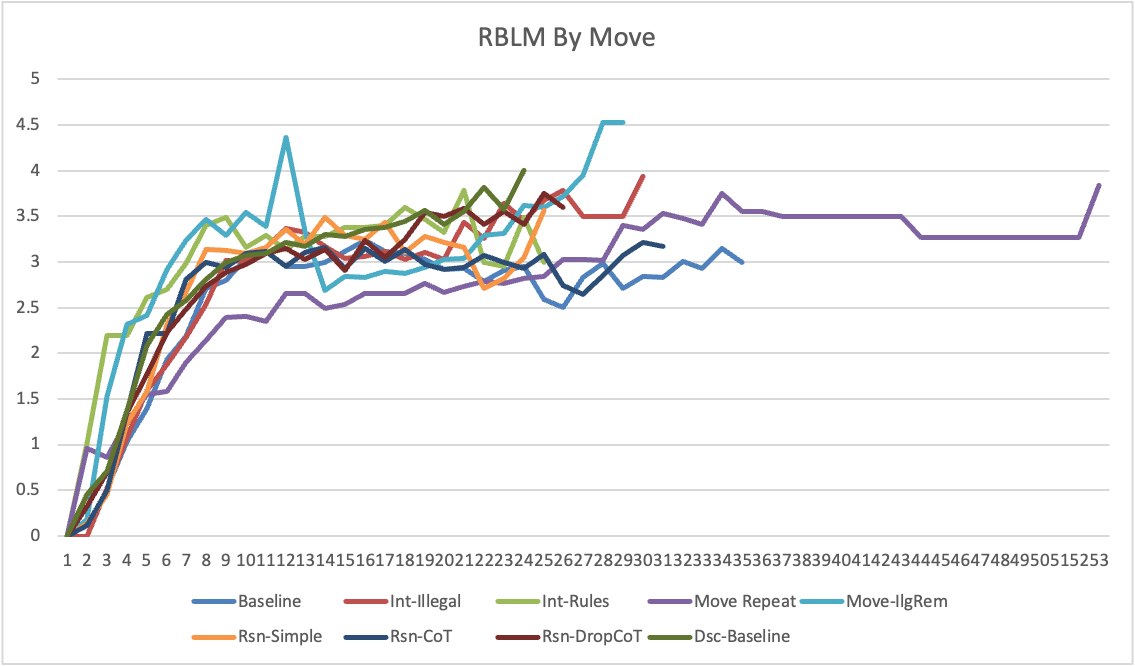}
    \caption{Average RBLM by Move}
    \label{fig:apd_rblm}
\end{figure}

\begin{figure}[H]
    \centering
    \includegraphics[width=7.2cm]{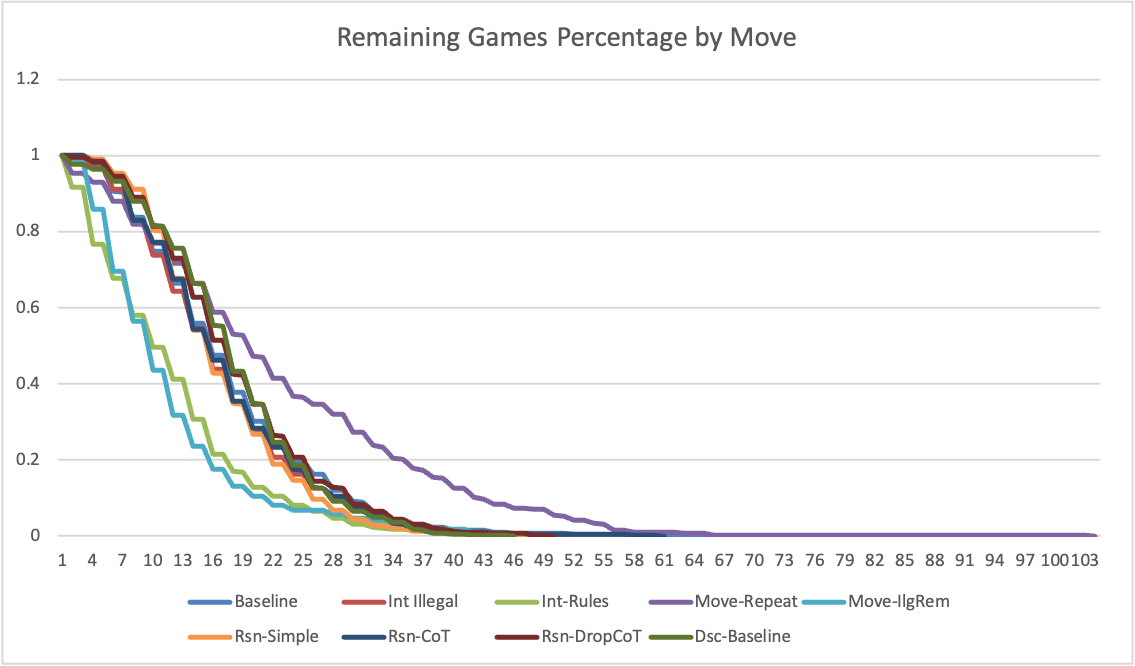}
    \caption{Remaining Games by Move}
    \label{fig:apd_rg}
\end{figure}

\begin{figure}[H]
    \centering
    \includegraphics[width=7.2cm]{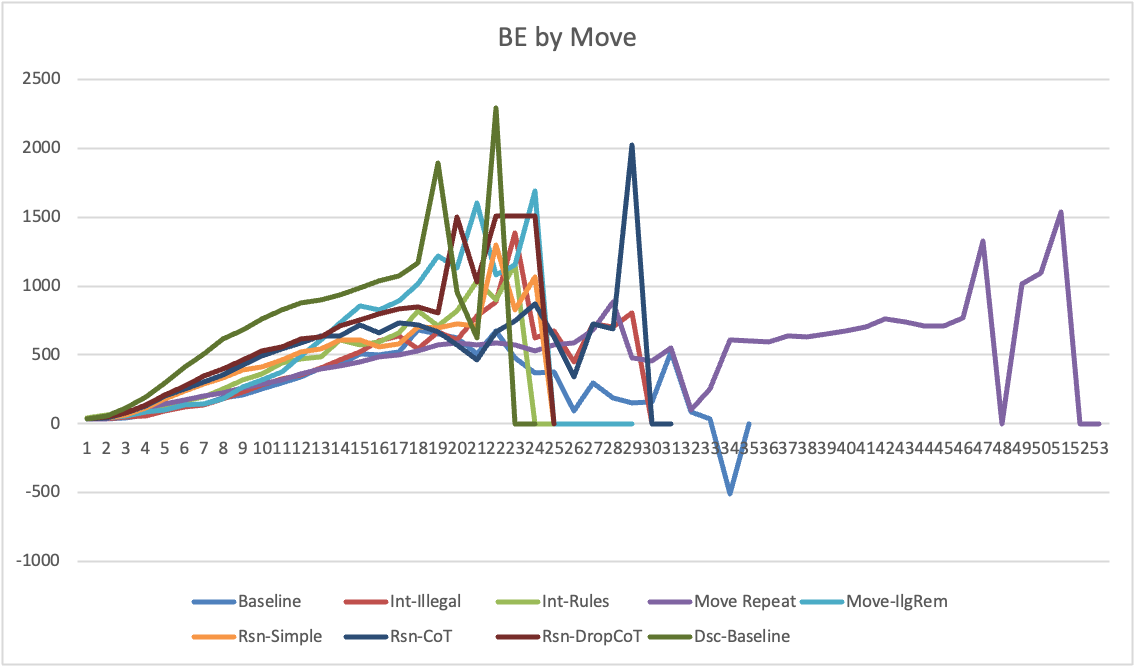}
    \caption{Board Evaluation by Move}
    \label{fig:apd_be}
\end{figure}

\end{document}